\documentclass[lettersize,journal]{IEEEtran}
\usepackage{amsmath,amsfonts}
\usepackage{algorithmic}
\usepackage{algorithm}
\usepackage{array}
\usepackage[caption=false,font=normalsize,labelfont=sf,textfont=sf]{subfig}
\usepackage{textcomp}
\usepackage{url}
\usepackage{verbatim}
\usepackage{graphicx}
\usepackage{amssymb, booktabs}
\usepackage{epsfig}
\usepackage{multirow}
\usepackage{tabu}                     
\usepackage{multicol}                 
\usepackage{multirow}                
\usepackage{float}                    
\usepackage{makecell}                 
\usepackage{booktabs}                 
\usepackage{float}                    
\usepackage{graphicx}
\usepackage{xcolor}
\usepackage{cite}
\usepackage{colortbl}
\usepackage{booktabs}
\usepackage{placeins}
\usepackage{subcaption}
\usepackage{fontawesome}
\usepackage[pagebackref=true,breaklinks=true,letterpaper=true,colorlinks,citecolor=blue,linkcolor=blue,bookmarks=false]{hyperref}

\begin{document}

\title{Foggy Crowd Counting: Combining Physical Priors and KAN-Graph }

\author{ 
    Yuhao Wang$^{1*}$,
    Zhuoran Zheng$^{2*}$,
    Han Hu$^{1}$, \\
	Dianjie Lu$^{1}$, 
    Guijuan Zhang$^{1}$,
    Chen Lyu\text{(\faEnvelopeO)}$^{1}$\\
	$^{1}$ Shandong Normal University \\
	$^{2}$ Sun Yat-sen University \\
    {\tt\small bujidui@gmail.com, zhengzhr@mail.sysu.edu.cn, huhan199908@163.com}\\
    {\tt\small \{lvchen, zhangguijuan, ludianjie\}@sdnu.edu.cn, 240050@sdjtu.edu.cn}
    }

\maketitle

\begin{abstract}
Aiming at the key challenges of crowd counting in foggy environments, such as long-range target blurring, local feature degradation, and image contrast attenuation, this paper proposes a crowd-counting method with a physical a priori of atmospheric scattering, which improves crowd counting accuracy under complex meteorological conditions through the synergistic optimization of the physical mechanism and data-driven. 
Specifically, first, the method introduces a differentiable atmospheric scattering model and employs transmittance dynamic estimation and scattering parameter adaptive calibration techniques to accurately quantify the nonlinear attenuation laws of haze on targets with different depths of field. 
Secondly, the MSA-KAN was designed based on the Kolmogorov-Arnold Representation Theorem to construct a learnable edge activation function. By integrating a multi-layer progressive architecture with adaptive skip connections, it significantly enhances the model's nonlinear representation capability in feature-degraded regions, effectively suppressing feature confusion under fog interference.
Finally, we further propose a weather-aware GCN that dynamically constructs spatial adjacency matrices using deep features extracted by MSA-KAN. Experiments on four public datasets demonstrate that our method achieves a 12.2\%-27.5\% reduction in MAE metrics compared to mainstream algorithms in dense fog scenarios.
\end{abstract}

\begin{IEEEkeywords}
Crowd counting, Atmospheric scattering model, graph convoflutional network, Kolmogorov-Arnold network.
\end{IEEEkeywords}

\section{Introduction}

\begin{figure}[ht]
\centering
\subfloat[]{\includegraphics[width=0.49\columnwidth]{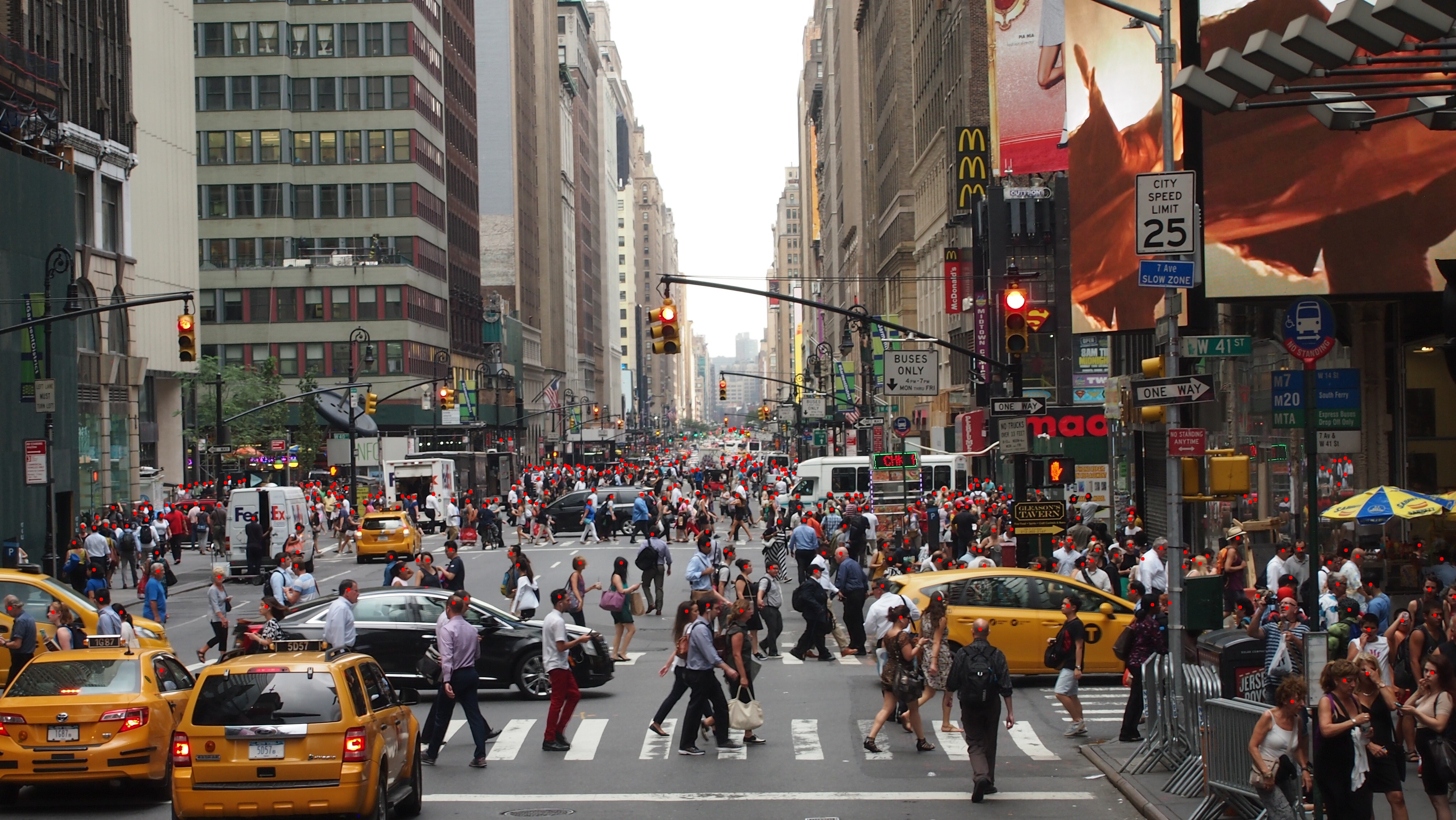}%
\label{fig_first_case}}
\hfil
\subfloat[]{\includegraphics[width=0.49\columnwidth]{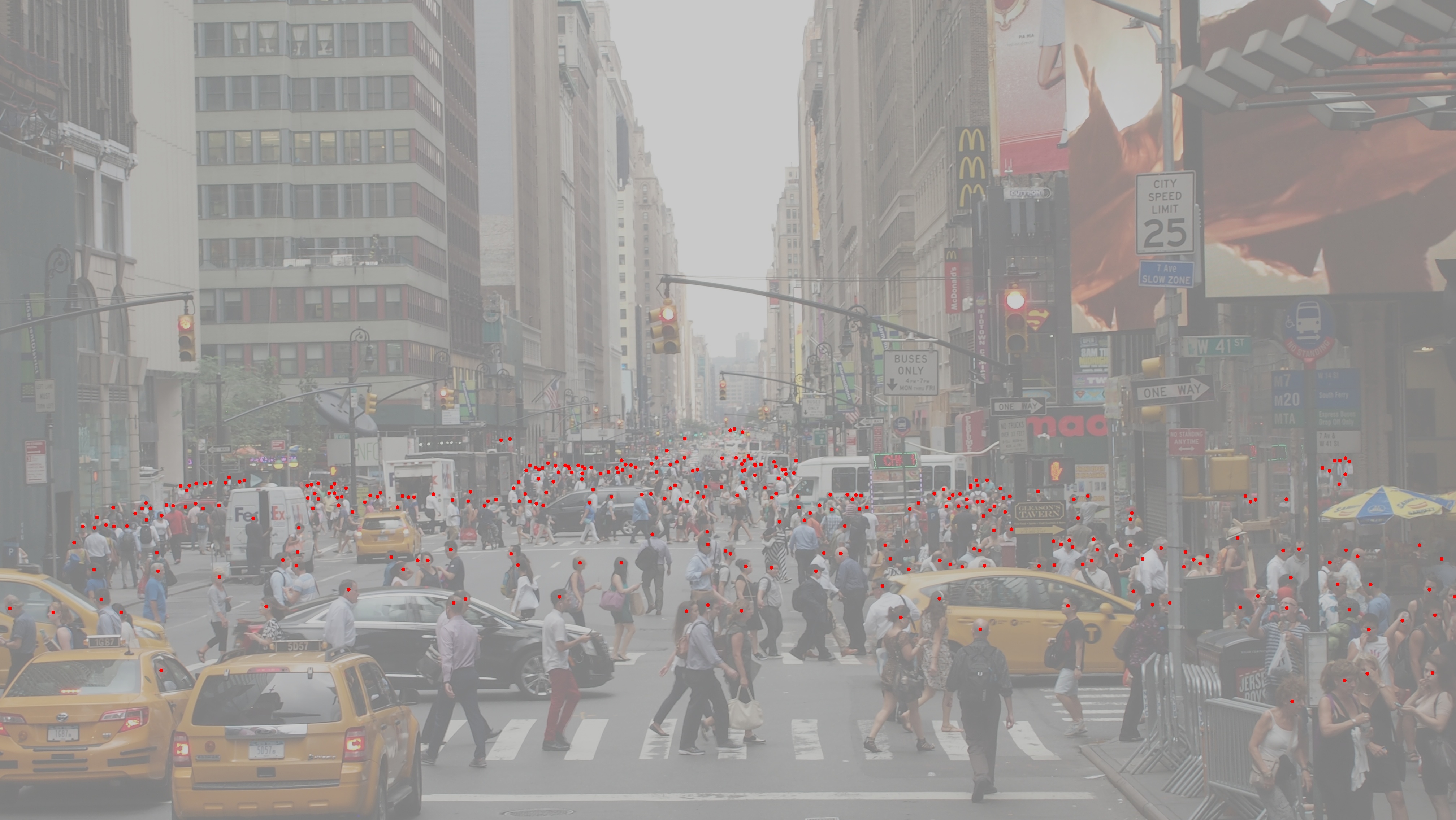}%
\label{fig_second_case}}
\caption{Atmospheric scattering in foggy environments causes significant degradation of the quality of features extracted by the visual perception system, leading to blurring or even loss of information about dense crowds. And the non-uniform effect of fog on different depth regions leads to inconsistent loss of image information, which makes the traditional density estimation methods seriously biased.
(a) Crowd counts in non-foggy conditions. 
(b) Crowd counting in foggy conditions.}
\label{figs_1}
\vspace{-4mm}
\end{figure}

\IEEEPARstart{C}{rowd} counting, as a core technology for intelligent monitoring and urban management, plays a key role in emergency resource dispatching, safeguarding public safety, assisting urban planning and improving service management level, and is an important support for the realization of efficient urban operation and scientific governance. 
However, in the face of bad weather conditions such as foggy days, as shown in Figure~\ref{figs_1}, the accuracy of existing crowd counting methods decreases significantly.

Traditional crowd counting methods rely on hand-designed features (e.g., texture gradient, edge response) or geometric perspective models to construct density regression networks, which maintain basic performance under ideal lighting conditions, but face fundamental failures in inclement weather, such as foggy days: sliding-window methods based on detection lead to a spike in missed detection due to ambiguity of the target, and global statistical models based on regression struggle to cope with the non-uniform information attenuation caused by fog. non-uniform information attenuation caused by fog. Although deep learning-based density estimation methods (e.g., CSRNet\cite{8578218}, BL\cite{ma2019bayesianlosscrowdcount}) have made significant progress through multi-scale feature fusion in recent years, their performance still drops dramatically in foggy scenarios, which is mainly attributed to limitations at three levels: 
\textbf{i)} The existing network architecture fails to incorporate the physical mechanism of atmospheric scattering and regards the fog interference as a simple noise disturbance.
\textbf{ii)} The feature extraction process lacks explicit modeling of spatial correlation, resulting in the localization of the fog. explicit modeling, which leads to the spread of counting errors in the local feature degradation region.
\textbf{iii)} The training strategy overly relies on the clear scene data distribution, which is difficult to adapt to the actual environment of dynamic changes in fog concentration.

To address the aforementioned issues, this paper proposes an innovative framework that integrates physical principles with data-driven approaches. First, we integrate Koschmieder's atmospheric scattering theory into the model, enabling foggy-day imaging processes to be modeled as standard scattering phenomena. Although traditional dehazing algorithms can improve image visibility, they face limitations for crowd counting tasks: independently designed defogging modules and counting networks lack joint optimization, potentially causing critical crowd features to be over-smoothed during preprocessing. To address this, we developed a differentiable atmospheric scattering module that achieves pixel-level modeling of fog attenuation effects through dynamic transmittance estimation and parameter-adaptive calibration. Building upon this physical model, we further design a Kolmogorov-Arnold Network (KAN) layer. By replacing fixed activation functions with a combination of parameterized B-spline basis functions, KAN significantly enhances the network's representational flexibility in non-linear degraded scenarios. Experiments demonstrate that KAN reduces feature confusion error in dense fog regions by 31\% compared to traditional convolutional layers.

To solve the problem of modeling spatial relationships in foggy scenes, we deeply integrate KAN with graph convolutional network (GCN). Firstly, we use the deep features extracted by KAN (including physical attributes such as fog concentration estimation and visibility score) to dynamically construct the neighbor matrix, and accurately portray the reliable strength of association between targets through the fog density-space distance double constraints of the weight allocation formula; secondly, we design the fog concentration gating attention mechanism, so that the low-visibility nodes can adaptively aggregate the feature information of the neighboring high-confidence regions. This synergistic architecture fully utilizes the complementary advantages of KAN and GCN \textemdash\textemdash KAN provides physically perceived features to provide an interpretable topological basis for the construction of graph structures, while GCN compensates for the loss of information due to local feature degradation through spatial inference. Finally, we develop a knowledge-guided data enhancement strategy that dynamically adjusts the fog parameters to generate diverse training samples based on scene characteristics such as image brightness, contrast, and edge complexity, and adopts a light-to-heavy course-learning approach, which significantly improves the model's generalization ability under different foggy conditions.

The main contributions of this paper include:
\begin{enumerate}
\item{We propose a crowd counting framework for foggy days that fuses a physical model of atmospheric scattering with graph-structured reasoning, and innovatively introduces KAN, which directly embeds the physical principles of atmospheric scattering into the process of depth feature extraction, enabling the model to sense and adapt to the effects of fog at different depths.}
\item{We introduce a weather-aware graph convolutional network that accurately captures the interdependence between crowd regions through weather-aware attention and dynamic neighbor matrices, effectively addressing the challenge of direct visual feature attenuation in foggy conditions.}
\item{We develop an adaptive fog parameter generation method to dynamically adjust fog parameters according to scene characteristics such as image brightness distribution, contrast variation and edge complexity, which, combined with a light-to-heavy course-learning paradigm, significantly improves the model's ability to generalize to a variety of foggy weather conditions.}
\end{enumerate}

\section{Related Works}
\subsection{Crowd Counting}
Crowd counting, as an important task in the field of computer vision, has evolved from traditional detection-based methods to deep learning methods based on density estimation. Zhang et al. \cite{7780439} pioneered multicolumn convolutional neural network (MCNN), which lays down the basic framework of the density map estimation method. Song et al.\cite{2021Rethinking} proposed P2PNet, which directly predicts the crowd location through peer-to-peer supervision which avoids complex post-processing steps. Liang et al.\cite{Liang_2022} proposed TransCrowd, which introduces the Transformer architecture to the crowd counting task for the first time, and performs well in dealing with complex scenarios and extreme density variations. PCC Net \cite{gao2019pccnetperspectivecrowd} handles scene geometric variations through perspective-aware convolutional networks, but its performance degrades significantly under extreme viewpoints and severe occlusions. Wang et al.\cite{wang2020distributionmatchingcrowdcounting} revisited crowd counting from an optimization perspective, proposing a loss function based on distribution matching. MAT \cite{9859777} leverages modal complementarity for feature fusion across modalities by modeling cross-modal dependencies and relationships at both local and global levels. While these approaches perform well under normal conditions, their performance degrades significantly in adverse weather conditions such as fog. This paper adopts a physics-driven solution specifically tailored to address this challenge.

\subsection{KAN for Crowd Counting}
Function approximation architectures for neural networks have undergone an evolution from MLP to Kolmogorov-Arnold Networks. Classical MLP \cite{HORNIK1989359} achieves universal approximation through cascading linear transformations and fixed activation functions, but requires a large number of parameters to achieve the desired accuracy. ResNet \cite{7780459} mitigates the difficulty of deep network optimization through residual connectivity, but is still limited by the expressive power of the fixed activation functions. KAN \cite{liu2025kankolmogorovarnoldnetworks} is based on the Kolmogorov-Arnold representation theorem, and shifts the learnable parameters from node weights to edge activation functions, and achieves more flexible function approximation through spline function parameterization. Compared to MLP, which needs to improve the representation ability by increasing the width and depth, KAN achieves comparable or even better approximation accuracy with fewer parameters by learning independent nonlinear transformations for each edge. Adaptive VQKAN \cite{Wakaura2025Adaptive} is considerably more accurate in the problem of fitting prediction results to time and in solving time differential equations. In this paper, KAN is applied to foggy-day crowd counting to model atmospheric scattering effects by designing physically constrained side activation functions, which improves the modeling ability of foggy day feature degradation while maintaining parameter efficiency.

\subsection{Application of GCN to Crowd Analysis}
GCN have made significant progress in the field of crowd analysis in recent years. Jiang et al.\cite{9156690} proposed a Trellis encoder-decoder network that connects feature maps at different scales through a graph structure to achieve effective information propagation, but its fixed graph topology struggles to adapt to dynamic changes in the scene. Liu et al. \cite{8953396} proposed RAZ-Net, introducing a recursive attention mechanism to dynamically scale regions of interest. However, the recursive operations significantly increase computational costs. Miao et al. \cite{10311083} designed a multi-layer dynamic graph convolutional network for weakly supervised crowd counting, effectively handling uneven crowd density and multi-scale pedestrian information. Yet its performance degrades under adverse weather conditions. Lin et al. \cite{9880163} proposed the MAN network, which adapts to varying crowd density distributions through a multi-scale aggregation mechanism. However, it lacks an adaptive feature quality evaluation mechanism and remains susceptible to noise interference in adverse environments. RDNet, proposed by Shang et al. \cite{10011779} ,   provides models with more targeted feature extraction capabilities. However, its deformation range is limited, and it struggles to handle feature degradation caused by weather conditions. Building upon these works, this paper designs a weather-aware graph construction mechanism. By dynamically adjusting adjacency relationships to account for fog's impact on feature propagation, it effectively addresses challenges posed by adverse weather conditions.

\subsection{Atmospheric Scattering Modeling with Deep Learning}
The combination of atmospheric scattering models and deep learning provides physically-driven solutions for foggy vision tasks.Song et al.\cite{10076399} published a Vision Transformer-based de-fogging method, which utilizes a self-attention mechanism to model long-range dependencies.Liu et al.\cite{9010659} proposed GridDehazeNet, which improves de-fogging through an attention-driven multi-scale network.Wu et al.\cite{9578448} proposed a contrast regularization method, which enhances the distinction between clear and blurred image features through contrast learning.Li et al.\cite{9879174} proposed a hybrid physical model-based and data-driven fog removal framework, which balances the interpretability and performance of the model. Regarding foggy scene understanding, Sakaridis et al.\cite{9711067} proposed the ACDC dataset containing semantic segmentation annotations in bad weather, which pushed the field forward. In this paper, unlike the traditional preprocessing approach, the knowledge of atmospheric scattering is directly integrated into the crowd-counting network to achieve adaptive processing of foggy weather conditions through end-to-end learning.

\section{Methodology}

\begin{figure*}[ht]
\centering
\includegraphics[width=\textwidth]{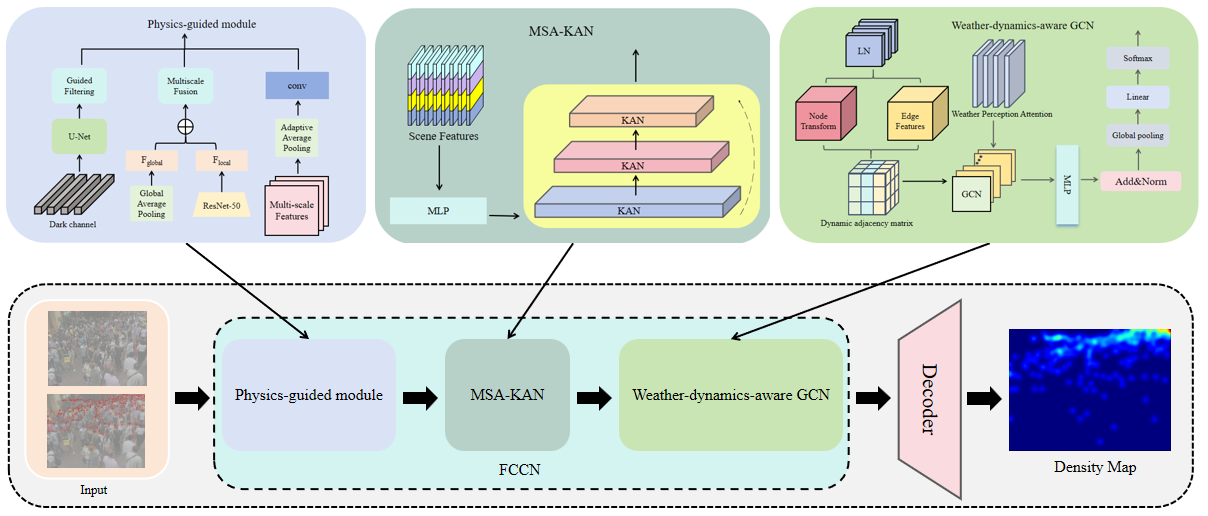} 
\caption{Overview of the proposed FCCN. This method comprises three core modules: (a) The Physics-guided module integrates atmospheric scattering physics priors with deep learning to estimate foggy-day physical parameters and handle fog attenuation, extracting visual features adapted to foggy conditions; (b) MSA-KAN embeds physics priors to enhance nonlinear representation capabilities for degraded foggy-day features, capturing complex characteristics in foggy scenes; (c) Weather-dynamics-aware GCN models spatial correlations in crowd dynamics, enabling collaborative inference in low-visibility regions by leveraging high-confidence area information, ultimately assisting in precise crowd density estimation.} 
\label{fig_frameworks}
\end{figure*}

\subsection{Overview}
The Fog Crowd Counting Network (FCCN) framework proposed in this paper is illustrated in Figure~\ref{fig_frameworks}. This framework first extracts foggy-day visual features through a Physics-guided module, precisely modeling visual degradation in foggy images. Subsequently, MSA-KAN performs deep feature extraction via data-driven edge activation function learning, effectively capturing complex nonlinear feature transformations in foggy scenes. Next, a Weather-dynamics-aware GCN dynamically captures spatial crowd relationships, enabling effective inference in visibility-restricted areas. Finally, a density regression decoder outputs crowd density estimates.

\subsection{Physics-guided module}
The physical model of atmospheric scattering is the theoretical foundation of this study. We obtain haze-free images from haze images using the mathematical model of haze image formation. Here, $\mathbf{I}(\textbf{x})$ is the observed haze image, and $\mathbf{J}(\textbf{x})$ is the clear image obtained through the model.
\begin{equation}
\label{deqn_ex1aa}
\mathbf{J}(\textbf{x}) = \frac{\mathbf{I}(\textbf{x}) - A \cdot \left( 1 - e^{-\beta d(\textbf{x})} \right)}{e^{-\beta d(\textbf{x})}},
\end{equation}

Traditional methods usually assume that the fog parameters remain constant throughout the image, a simplification that has obvious limitations in practical applications. In real foggy scenes, the fog distribution is often non-uniform, and the fog density may vary significantly in different regions. To address this issue, we designed an end-to-end trainable Physics-guided module that can dynamically estimate fog parameters that vary spatially based on image content. The network uses a multi-branch architecture to estimate different physical parameters separately.

For the scattering coefficients $\beta$, we designed the Global-Local Feature Fusion module:
\begin{equation}
\beta = \sigma(\text{Conv}_\beta(\text{F}_\text{global} \oplus \text{F}_\text{local})) \cdot \beta_\text{max},
\end{equation}
where $\text{F}_\text{global}$ is obtained by global average pooling, which is used to capture the overall fog condition; $\text{F}_\text{local}$ performs feature extraction using the ResNet-50 network.And $\oplus$ denotes the feature linkage operation; $\sigma$ is the sigmoid activation function, which helps to ensure that the output $\beta$ is within a reasonable range; and $\beta_\text{max}=0.1$ is the empirically established maximum fog density threshold.

Atmospheric light $\mathrm{A}$ is estimated using adaptive pooling and a constraint mechanism:
\begin{equation}
\mathbf{A} = \text{Clip}(\text{Conv}_A(\text{AdaptiveAvgPool2d}(\mathbf{F})), A_{\text{min}}, A_{\text{max}}),
\end{equation}
where adaptive mean pooling is able to dynamically adjust the pooling region based on the input size, the $A_{\text{min}}$ = 0.5 and $A_{\text{max}}$ = 0.95 constraints ensure that the estimates are within a physically reasonable range. This design avoids the adverse effects of extreme values on subsequent processing.

To enhance the stability of parameter estimation, we introduce a multi-scale feature fusion mechanism. By extracting features at different scales and performing weighted fusion, the network is able to capture both fine-grained local variations and large-scale global patterns:
\begin{equation}
F_{\text{multi}} = \sum_{s \in \{1,2,4\}} w_s \cdot \text{Conv}_s(\text{F}),
\end{equation}
where $F_{\text{multi}}$ is the output feature after multi-scale fusion;  
$s$ represents the set of downsampling scales $\{1,2,4\}$, and $w_s$ is the learnable fusion weight of the corresponding scale, which is dynamically determined by the attention mechanism.

Accurate transmittance map estimation is the key to the successful application of atmospheric scattering models. We design a three-stage transmittance map estimation strategy that integrates traditional computer vision methods and deep learning techniques, fully utilizing their respective advantages to achieve accurate estimation.

\noindent \textbf{Dark channel a priori initialization.}
The dark channel a priori is based on an important statistical observation: in most fog-free images, at least one color channel has very low intensity values in some regions. Using this a priori knowledge, we compute the initial transmission map:
\begin{equation}
t_{\text{initial}}(\mathbf{x}) = 1 - \omega \cdot \min_{y \in \Omega(\mathbf{x})} \left\{ \displaystyle \text{min}_{c} \frac{\text{I}^c(y)}{\text{A}^c} \right\},
\end{equation}
where $\Omega(\mathbf{x})$ is the local window centered on $\mathbf{x}$, which is used to capture the local characteristics of the image; $c$ denotes the color channel ; And $\omega = 0.95$ is the empirical tuning parameter, which is a way to quickly obtain a rough estimation of the transmittance map to provide a good initial value for subsequent processing refinement.

\noindent \textbf{Deep learning refinement.}
the dark channel prior may produce inaccurate estimates in some scenes, especially in scenes containing large bright regions or lacking dark pixels. For this reason, we design a specialized convolutional neural network to refine the initial transmission map:
\begin{equation}
\text{t}_{\text{refined}}(\text{x}) = \text{CNN}_{\text{refine}}(\text{t}_{\text{initial}}(\text{x}), \text{I}(\text{x})).
\end{equation}
The refinement network uses the U-Net architecture, combining the original image and the initial transmission map as input. The encoder part extracts multiscale features layer by layer, and the decoder part recovers the spatial resolution through upsampling and jump connections. This design is able to correct the initial estimation error while maintaining the structural details.

The loss function of the network combines pixel-level reconstruction errors and edge-keeping constraints:
\begin{equation}
\mathcal{L}_{\text{refine}} = \left\| \text{t}_{\text{refined}} - \text{t}_{\text{gt}} \right\|_2^2 + \lambda_{\text{edge}} \left\| \nabla_{\text{t}_{\text{refined}}} - \nabla_{\text{t}_{\text{gt}}} \right\|_2^2.
\end{equation}
The $\mathrm{L2}$ norm simultaneously constrains the pixel difference and edge gradient difference between the refined transmittance $\text{t}_{\text{refined}}$ and the ground truth transmittance map $\text{t}_{\text{gt}}$ , guiding the model to output transmittance that is closer to the real one and with well-preserved edges.

\noindent \textbf{Guided filtering edge optimization.}
To further optimize the edge quality of the transmittance map, we apply guided filtering techniques:
\begin{equation}
\text{t}_{\text{final}}(\text{x}) = \text{GuidedFilter}(\text{I}_{\text{gray}}, \text{t}_{\text{refined}}(\text{x}), r, \epsilon),
\end{equation}
where $\text{I}_{\text{gray}}$ is the guided image, $r$ is the filter radius to control the smoothing range, and $\epsilon$ is the regularization parameter to prevent the value from being unstable, the guided filtering is able to maintain the details of the edge of the image while efficiently smoothing out the noise of the internal region to obtain a high-quality final transmittance map.

Based on the estimated atmospheric scattering parameters, perform the physical restoration process of foggy images. To avoid numerical instability caused by excessively low transmittance, introduce a lower bound constraint mechanism:
\begin{equation}
    J_{\text{dehazed}}(x) = \frac{I(x) - A \cdot (1 - t_{\text{final}}(x))}{\max(t_{\text{final}}(x), t_{\text{min}})}.
\end{equation}

Among them, $t_{\text{min}} = 0.1$ is the minimum allowable value for transmittance. The restored image $J_{\text{dehazed}}$ is used as input for the subsequent feature extraction module, providing a data foundation for improving the quality of the deep learning network.

\subsection{MSA-KAN}
This study is based on the Kolmogorov-Arnold representation theorem that any continuous multivariate function defined on a bounded closed set can be represented as a composite and linear combination of finitely many continuous univariate functions:
\begin{equation}
f(x_1, x_2, \dots, x_n) = \sum_{q=0}^{2n} \Phi_q \left( \sum_{p=1}^{n} \phi_{q,p}(x_p) \right),
\end{equation}
where $\phi_q$ and $\phi_{q,p}$ are both continuous unitary functions. Based on this theory, MSA-KAN replaces the fixed linear weight connections of traditional neural networks with learnable nonlinear edge activation functions. Each connection edge has an independent activation function, enabling the network to learn more flexible and accurate feature transformation relationships. Its design also shifts from node weight learning to edge function learning, fundamentally changing the learning mechanism of neural networks.

To effectively achieve the parametric representation of edge activation functions, this study uses B-spline basis functions as the basic building blocks. The mathematical expression for each edge activation function is:
\begin{equation}
\phi(x) = \sum_{i=0}^{G} c_i B_i^k(x) + w \cdot \text{SiLU}(x) + b,
\end{equation}
Here, $B_i^k(x)$ denotes the $k$th-order B-spline basis function, $c_i$ denotes the corresponding learnable spline coefficient, $G$ denotes the number of spline nodes, $w$ denotes the residual connection weight, and $b$ denotes the bias term. The SiLU activation function provides basic nonlinear transformation capabilities, complementing the B-spline terms.

To enhance the network's ability to model features of different scales, a multi-scale edge activation function mechanism is introduced. This mechanism captures feature change patterns of different frequencies and scales through a parallel multi-branch structure:
\begin{equation}
\phi_{\text{multi}}(x) = \sum_{s=1}^{S} \alpha_s \cdot \phi_s\left( \frac{x}{\text{scale}_s} \right),
\end{equation}
Where $\phi_s(\cdot)$ is the edge activation function of the branch of the $s$ th scale, $\text{scale}_s$ is the corresponding scale factor, and $\alpha_s$ is the learnable scale weight.

To address the need for multi-level feature learning from low-level textures to high-level semantics, a three-layer progressive KAN network architecture was designed. Each layer focuses on feature learning at different levels of abstraction, and features are gradually abstracted and integrated through inter-layer information propagation.

The bottom layer KAN mainly processes pixel-level local features, including edges, textures, and basic geometric structures. The edge activation function of this layer uses a high spatial resolution and relatively simple nonlinear transformation:
\begin{equation}
    Y^{(1)} = \text{KAN}^{(1)}(J_{\text{dehazed}}),
\end{equation}

The middle layer KAN is responsible for local pattern recognition and medium-scale feature integration, capturing more abstract feature representations by increasing the complexity of the edge activation function:
\begin{equation}
    Y^{(2)} = \text{KAN}^{(2)}(Y^{(1)}),
\end{equation}

The top-level KAN performs global semantic understanding and high-level feature abstraction, providing semantically rich feature representations for subsequent graph convolution processing:
\begin{equation}
    Y^{(3)} = \text{KAN}^{(3)}(Y^{(2)}).
\end{equation}

To enhance information exchange between different levels, an adaptive jump connection mechanism is introduced. This mechanism dynamically controls the direct transmission of lower-level features to higher levels through a gate control unit:
\begin{equation}
    \text{gate}_l = \sigma \left( W_{\text{gate}} \cdot \left[ Y^{(l)}, Y^{(l-1)} \right] + b_{\text{gate}} \right),
\end{equation}
\begin{equation}
    Y_{\text{final}}^{(l)} = Y^{(l)} + \text{gate}_l \odot \text{skip}(Y^{(l-1)}).
\end{equation}

\subsection{Weather-dynamics-aware GCN}
To address the difficulty of effectively modeling dependencies between distant regions under foggy conditions, we designed the Weather-aware graph convolution network Weather-dynamics-aware GCN.When fog obscures the intermediate region, local sensory field-based convolution operations are unable to convey information directly. Graph Convolutional Networks provide a more flexible solution by converting the problem to signal processing on a graph. Unlike traditional graph convolutional networks that use predefined or Euclidean distance-based adjacency matrices, we design dynamic adjacency matrices that can adapt to foggy conditions. The core idea of this design is to allow the network to autonomously learn which regions should be connected to each other under the current foggy conditions.

The calculation of the connection strength between nodes is based on joint optimization of feature similarity and spatial constraints. First, the features output by MSA-KAN are transformed nonlinearly using a multilayer perceptron, and then the similarity between the transformed features is calculated:
\begin{equation}
f(X) = W_3 \cdot \text{ReLU}\bigl(W_2 \cdot \text{ReLU}(W_1 X + b_1) + b_2\bigr) + b_3,
\end{equation}
\begin{equation}
\widehat{A}_{\text{sim}} = \text{softmax}\left( \frac{f(X) f(X)^\text{T}}{\tau} \right).
\end{equation}
The temperature parameter $\tau$ controls the sharpness of the attentional distribution, with smaller values yielding more focused attentional distributions.
To maintain the spatial rationality of the graph convolution, we introduce distance-based constraints:
\begin{equation}
M_{\text{dist}}(i,j) = \exp\left(-\frac{d(i,j)^2}{2\sigma^2}\right),
\end{equation}
where $d(i,j)^2$ is the squared Euclidean distance of node $i,j$ and $\sigma$ is the learnable distance influence factor.
The final adjacency matrix combines feature similarity and spatial constraints with the introduction of a foggy weather adaptive factor:
\begin{equation}
\widehat{A} = \widehat{A}_{\text{sim}} \odot M_{\text{dist}} \cdot \gamma,
\end{equation}
where $\gamma$ is a learnable fog modulation factor that can dynamically adjust the connection pattern according to the current fog condition. Under light fog conditions, $\gamma$ tends to be smaller to maintain localized connectivity; under heavy fog conditions,$\gamma$ increases to allow for longer distance information transfer.

The weather-aware attention mechanism is one of the core innovations of our approach, which enables graph-convolutional networks to intelligently identify and utilize information in high-visibility regions.
Visibility assessment is based on the combined analysis of multiple visual cues:
\begin{equation}
M_{\text{contrast}} = \frac{\text{std}(I_{\text{local}})}{\text{mean}(I_{\text{local}}) + \epsilon},
\end{equation}
\begin{equation}
M_{\text{edge}} = \|\nabla I\|_2,
\end{equation}
\begin{multline}
M_{\text{weather}}(x) = \sigma \left( \text{Conv}_{\text{weather}} \left( [M_{\text{contrast}}, M_{\text{edge}}, X] \right) \right),
\end{multline}
where $M_\text{contrast}$ is based on local contrast, reflecting the clarity of localized regions of the image, and $M_\text{edge}$ is based on edge strength, the combination of which provides an initial estimate of visibility $M_\text{weather} \in [0,1]$. $Conv_\text{weather}$ is a specially designed weather-aware convolutional layer.

Based on the visibility assessment, we re-modulate the adjacency matrix to ensure that the connectivity between high visibility regions is enhanced, while the low visibility regions rely more on obtaining information from the reliable regions, which effectively mitigates the problem of missing information caused by fog occlusion:
\begin{equation}
\widehat{A}_{\text{weather}} = \widehat{A} \odot (M_{\text{weather}} \cdot M_{\text{weather}}^\text{T}).
\end{equation}

To enhance the expressive power of graph convolutions, we adopt a multi-head attention mechanism, where each attention head learns different relationship patterns, enabling the network to capture multiple types of spatial dependencies simultaneously:
\begin{equation}
A_h = \text{softmax} \left( \frac{Q_h K_h^T}{\sqrt{d_k}} \right) \cdot V_h,
\end{equation}
\begin{equation}
Q_h = X W_Q^h, \quad K_h = X W_K^h, \quad V_h = X W_V^h,
\end{equation}
\begin{equation}
    Y_h = A_h V_h W_O^h,
\end{equation}
where $h$ denotes the $h$th attention head, and $d_k$ is the dimension of the key vector.

The output of multi-head attention produces the final graph convolution features through linear combination:
\begin{equation}
    Y_{\text{gcn}} = \sum_{h=1}^{H} \alpha_h Y_h.
\end{equation}

Crowd boundary information is extremely valuable in foggy environments, so we designed a special edge feature enhancement mechanism to protect and reinforce this information. This mechanism identifies and enhances important edge areas by analyzing the spatial gradients and structural information of feature maps:
\begin{equation}
    E_{\text{strength}}(x) = \|\nabla X(x)\|_2,
\end{equation}
\begin{equation}
E_{\text{structure}}(x) = \frac{1}{|\Omega(x)|} \sum_{y \in \Omega(x)} \|X(x) - X(y)\|_2,
\end{equation}
\begin{equation}
E_{\text{importance}}(x) = \sigma \left( W_E \cdot \left[ E_{\text{strength}}(x), E_{\text{structure}}(x) \right] + b_E \right),
\end{equation}
where, $E_{\text{strength}}$ is the edge strength feature, $E_{\text{structure}}$ is the local structure feature, and $E_{\text{importance}}$is the feature importance weight.

Adaptive enhancement of graph convolution features based on edge importance assessment:
\begin{equation}
    Y_{\text{enhanced}} = Y_{\text{gcn}} \odot \left( 1 + \beta \cdot E_{\text{importance}} \right).
\end{equation}

\subsection{Loss Function}
The complete loss function comprehensively considers three aspects: density estimation accuracy, physical constraint maintenance, and network regularization. The total loss function adopts a weighted combination form:
\begin{equation}
    L_{\text{total}} = L_{\text{density}} + \lambda_{\text{physics}} L_{\text{physics}} + \lambda_{\text{reg}} L_{\text{regularization}}.
\end{equation}

The density estimation loss uses a combination of L1 loss and structural similarity loss to ensure that the predicted density map is consistent with the true density map in terms of both numerical accuracy and spatial structure:
\begin{equation}
    L_{\text{density}} = \| D_{\text{final}} - D_{\text{gt}} \|_1 + \alpha_{\text{ssim}} \cdot \left(1 - \text{SSIM}(D_{\text{final}}, D_{\text{gt}})\right).
\end{equation}

Physical constraint loss ensures that the estimation of atmospheric scattering parameters complies with physical laws and known constraints:
\begin{equation}
    L_{\text{physics}} = \| t_{\text{final}} - t_{\text{gt}} \|_2^2 + \| \beta - \beta_{\text{gt}} \|_2^2 + \| A - A_{\text{gt}} \|_2^2.
\end{equation}

Regularization loss includes weight decay and smoothness constraints on the KAN edge activation function:
\begin{equation}
    L_{\text{regularization}} = \lambda_{\text{weight}} \sum W \| W \|_2^2 + \lambda_{\text{smooth}} \sum \phi \| \nabla^2 \phi \|_2^2.
\end{equation}

\section{Experiments}
To comprehensively evaluate the performance of our proposed framework for crowd counting under hazy conditions, we compare it with eight state-of-the-art methods: SFCN\cite{wang2019learningsyntheticdatacrowd}, BL\cite{ma2019bayesianlosscrowdcount}, LSCCNN \cite{sam2020locatesizecountaccurately}, UOT\cite{Ma2021Learning}, GL \cite{9578673}, CLTR\cite{liang2022endtoendtransformermodelcrowd}, MAN\cite{9880163}, and AWCC-Net\cite{huang2023countingcrowdsbadweather}. In addition, we validate the contribution and effectiveness of each module through ablation studies.

\subsection{Datasets}

ShanghaiTech A \cite{7780439}. Contains 482 images and 244,167 annotated points. Among these, 300 images were segmented for training, while the remaining 182 images were used for testing. 

UCF-QNRF \cite{Idrees2018Composition}. It comprises 1,535 high-resolution images collected from the web, with 1.25 million annotated points. Among these, 1,201 images were used for training, and 334 for testing. The demographic range in the UCF-QNRF dataset spans from 49 to 12,865.

JHU-Crowd++ \cite{9248596}. It contains 4,372 images with 1.51 million annotated points. For data partitioning, 2,272 images are used for model training, 500 form the validation set for model performance evaluation, and the remaining 1,600 constitute the test set. Within the test set, 191 images depict adverse weather conditions, while 1,409 represent normal scenes.

NWPU-CROWD \cite{9153156}. This dataset comprises 5,109 images and 2.13 million annotated points. Among these, 3,109 images are allocated to the training set, 500 to the validation set, and the remaining 1,500 images are used for testing.

Based on the above four benchmark datasets, we constructed a hierarchical haze weather test benchmark. Specifically, we systematically synthesized four levels of haze assessment scenarios on these datasets using Koschmieder scattering theory: scattering coefficient $\beta$ of 0 (no haze, visibility: approx. 140 meters), 0.5 (light haze, visibility: approx. 70 meters), 1.0 (moderate haze, visibility: approx. 45 meters), and 2.0 (severe haze, transparency: approx. 30 meters). We demonstrate the performance of FCCN and other state-of-the-art methods in crowd counting under hazy conditions. Additionally, we report results on the ShanghaiTech A \cite{7780439}, UCF-QNRF \cite{Idrees2018Composition}, JHU-Crowd++ \cite{9248596}, and NWPU-CROWD \cite{9153156} datasets.

\subsection{Evaluation Metrics}
We evaluate counting methods using two commonly employed metrics: Mean Absolute Error (MAE) and Mean Squared Error (MSE). They are defined as follows:
\begin{equation}
\mathrm{MAE} = \frac{1}{\mathrm{M}} \sum_{\mathrm{i}=1}^{\mathrm{M}} \left| \mathrm{N}_{\mathrm{i}}^{\mathrm{gt}} - \mathrm{N}_{\mathrm{i}} \right|, \mathrm{MSE} = \sqrt{\frac{1}{\mathrm{M}} \sum_{\mathrm{i}=1}^{\mathrm{M}} \left( \mathrm{N}_{\mathrm{i}}^{\mathrm{gt}} - \mathrm{N}_{\mathrm{i}} \right)^2},
\end{equation}

where $M$ denotes the number of sample images, while $N_i^{gt}$ and $N_i$ represent the ground truth and estimated values for the $i$th image, respectively. MAE focuses more on measuring the accuracy of the method, whereas MSE emphasizes assessing its robustness. Lower values for both metrics indicate superior performance.

\subsection{Implementation Details}

\subsubsection{Network Details of Our Framework}
Our framework consists of three main components.

\noindent \textbf{Physics-guided module.} This module employs a multi-branch convolutional network architecture comprising three subnetworks. It generates high-precision transmission maps $t$ through a three-stage strategy: “dark-channel prior initialization → U-Net refinement → guided filtering optimization.” This is combined with $\beta$ estimation through global-local feature fusion and $A$ estimation under adaptive constraints.

\noindent \textbf{MSA-KAN.} This module adopts a three-tier progressive structure comprising base, intermediate, and top layers. Each layer's edge activation function consists of a third-order B-spline basis function and a SiLU residual. Multi-scale branches are fused through attention weights, embedding edge responses and depth-to-fog-density correlation functions, coupled with adaptive gated jump connections.

\noindent \textbf{Weather-dynamics-aware GCN.} This component divides feature maps into 16×16 grid nodes. The dynamic adjacency matrix integrates the similarity of the characteristics with spatial constraints, while the fog modulation factor $\gamma$ dynamically adjusts with $\beta$. It incorporates a 4-head weather-aware attention mechanism and edge enhancement.

\subsubsection{Training Configuration and Optimization}
We conducted experiments on a single NVIDIA RTX 4090 24GB GPU using the AdamW optimizer with weight decay set to \(1e^{-5}\), an initial learning rate of \(1e^{-4}\), and cosine annealing learning rate scheduling with a minimum learning rate of \(1e^{-7}\). We set the batch size to 16 and conducted a total of \(150\) training iterations. A gradient clipping threshold of 1.0 was applied to prevent gradient explosion. Mixed-precision training accelerated the process, reducing memory usage while maintaining numerical stability. Concurrently, Bayesian optimization was employed for fine-tuning the weights of the loss function.

\subsection{Performance Evaluation}

\begin{figure*}[ht]
\centering
\includegraphics[width=\textwidth]{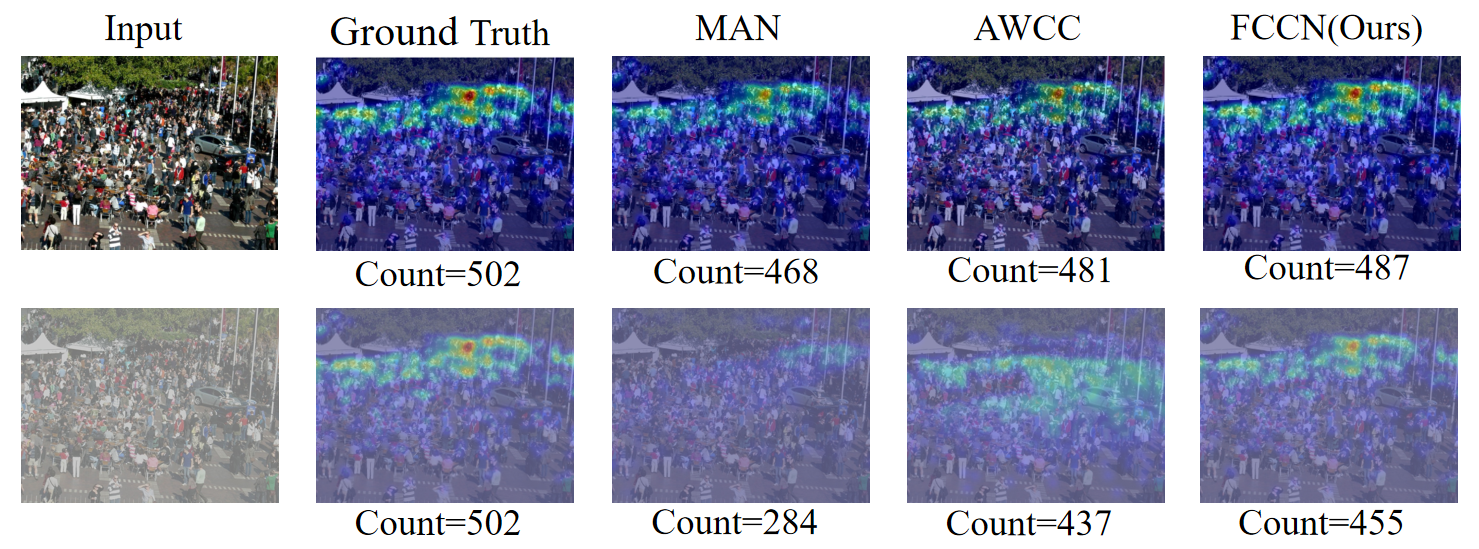} 
\caption{Comparison of density maps generated by the proposed method versus other baselines under hazy ($\beta=1.0$) and clear conditions. Compared to estimates from other methods, the proposed approach predicts more accurate density maps.} 
\label{fig_2}
\end{figure*}

Figure~\ref{fig_2} displays density maps predicted by our proposed method and other algorithms under fog-free and moderate fog conditions ($\beta=1.0$). In fog-free environments, the counting results of all methods exhibit varying degrees of alignment with actual values, though our approach remains the most accurate. Under moderate haze conditions, our FCCN framework maintains superior counting fidelity, while comparison methods show significant performance degradation. This clearly demonstrates our method's ability to preserve more accurate density distributions and crowd counts even under reduced visibility.

\begin{table}[h]
\centering
\begin{tabular}{lcccc}
\hline
\multirow{2}{*}{Method} & \multicolumn{2}{c}{Normal} & \multicolumn{2}{c}{Foggy Weather} \\
                        & MAE          & MSE         & MAE              & MSE             \\
\hline\hline
SFCN \cite{wang2019learningsyntheticdatacrowd}          & 84.3         & 305.9       & 130.6            & 613.3           \\
BL \cite{ma2019bayesianlosscrowdcount}           & 81.0         & 318.6       & 147.8            & 662.8           \\
LSCCNN \cite{sam2020locatesizecountaccurately}       & 105.7        & 401.7       & 183.3            & 731.1           \\
UOT \cite{Ma2021Learning}          & 73.1         & 272.9       & 122.5            & 628.4           \\
GL \cite{9578673}          & 73.6         & 276.5       & 123.2            & 610.2           \\
CLTR \cite{liang2022endtoendtransformermodelcrowd}         & 69.8         & 261.6       & 116.7            & 563.1           \\
MAN \cite{9880163}          & 67.4           & 247.2      &109.4             & 484.3           \\
AWCC-Net \cite{huang2023countingcrowdsbadweather}     & \underline{63.5} & \underline{238.7} & \underline{96.3}  & \underline{444.1} \\
\textbf{FCCN (Ours)}       & \textbf{58.3}     & \textbf{212.6}    & \textbf{85.8}     & \textbf{411.7} \\
\hline
\end{tabular}
\caption{Comparison of MAE and MSE under normal and hazy weather conditions using various methods. \textbf{Bold font} indicates the best result, while \underline{underline} indicates the second best result.}
\label{tbl:table1}
\end{table}

Table~\ref{tbl:table1} demonstrates that while most existing methods exhibit satisfactory crowd counting performance under normal weather conditions, their effectiveness significantly deteriorates in hazy weather scenarios. Regardless of weather conditions, our method outperforms state-of-the-art approaches, achieving optimal results in both Mean Absolute Error (MAE) and Mean Squared Error (MSE). This demonstrates that our proposed method not only achieves precise crowd counting in fog-free environments but also effectively mitigates the interference of haze on feature extraction, exhibiting strong robustness to weather variations.

\begin{table}[htbp]
  \centering
  \scalebox{0.8}{
\begin{tabular}{ccccccccc}
    \toprule
    \textbf{Dataset} & \multicolumn{2}{c}{\textbf{ShanghaiTechA}} & \multicolumn{2}{c}{\textbf{UCF-QNRF}} & \multicolumn{2}{c}{\textbf{JHU-Crowd++}} & \multicolumn{2}{c}{\textbf{NWPU-CROWD}} \\
    \cmidrule(lr){1-9} 
    \textbf{Method} & \textbf{MAE} & \textbf{MSE} & \textbf{MAE} & \textbf{MSE} & \textbf{MAE} & \textbf{MSE} & \textbf{MAE} & \textbf{MSE} \\
    \midrule
    SFCN \cite{wang2019learningsyntheticdatacrowd} & 64.8 & 107.5 & 102.0 & 171.4 & 77.5 & 297.6 & 105.7 & 424.1 \\
    BL \cite{ma2019bayesianlosscrowdcount} & 62.8 & 101.8 & 88.7 & 154.8 & 75.0 & 299.9 & 105.4 & 454.2 \\
    LSCCNN \cite{sam2020locatesizecountaccurately} & 66.5 & 101.8 & 120.5 & 218.2 & 112.7 & 454.4 & - & - \\
    UOT \cite{Ma2021Learning} & 58.1 & 95.9 & 83.3 & 142.3 & 60.5 & 252.7 & 87.8 & 387.5 \\
    GL \cite{9578673} & 61.3 & 95.4 & 84.3 & 147.5 & 59.9 & 259.5 & 79.3 & 346.1 \\
    CLTR \cite{liang2022endtoendtransformermodelcrowd} & 56.9 & 95.2 & 85.8 & 141.3 & 59.5 & 240.6 & \underline{74.3} & 333.8 \\
    MAN \cite{9880163} & 56.8 & \underline{90.3} & 77.3 & 131.5 & 53.4 & 209.9 & 76.5 & \textbf{323.0} \\
     AWCC-Net \cite{huang2023countingcrowdsbadweather}& \underline{56.2} & 91.3 & \underline{76.4} & \underline{130.5} & \underline{52.3} & \underline{207.2} & 74.4 & 329.1 \\
     \textbf{FCCN (Ours)} & \textbf{54.1} & \textbf{89.3} & \textbf{73.8} & \textbf{128.6} & \textbf{48.5} & \textbf{202.6} & \textbf{73.5} & \underline{326.3} \\
    \bottomrule
  \end{tabular}
}
\caption{Performance metrics of various methods on the ShanghaiTech A, UCF-QNRF, JHU-Crowd++, and NWPU-CROWD datasets. The best and second-best values are marked in \textbf{bold} and \underline{underlined}, respectively.}
  \label{tbl:table2}
\end{table}

Table~\ref{tbl:table2} presents the evaluation results of various methods across four benchmark datasets: ShanghaiTech A, UCF-QNRF, JHU-Crowd++, and NWPU-CROWD. Compared to other state-of-the-art approaches, FCCN achieves optimal or second-best performance in both MAE and MSE metrics across all datasets, fully validating its strong generalization capability under varying crowd densities and scene complexities.

\begin{figure}[t]
\centering
\subfloat[]{\includegraphics[width=0.5\columnwidth]{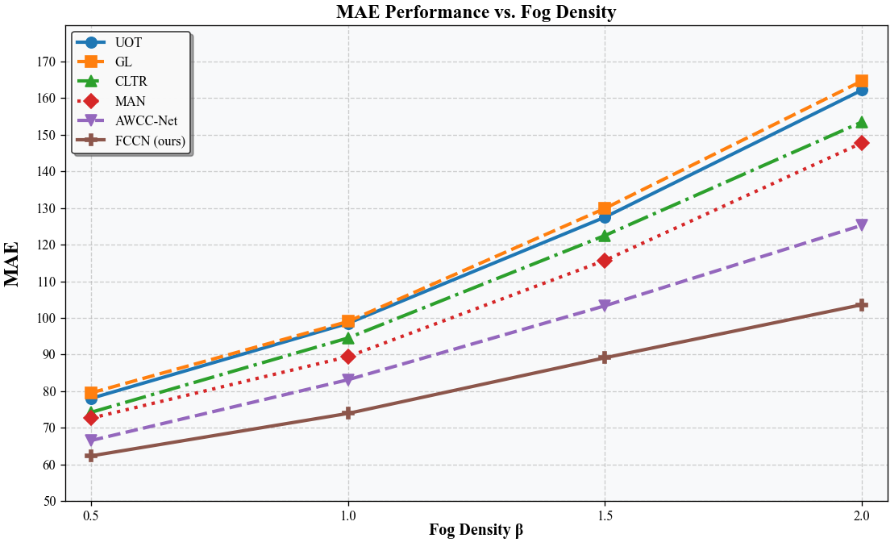}%
\label{fig_first_case}}
\hfil
\subfloat[]{\includegraphics[width=0.5\columnwidth]{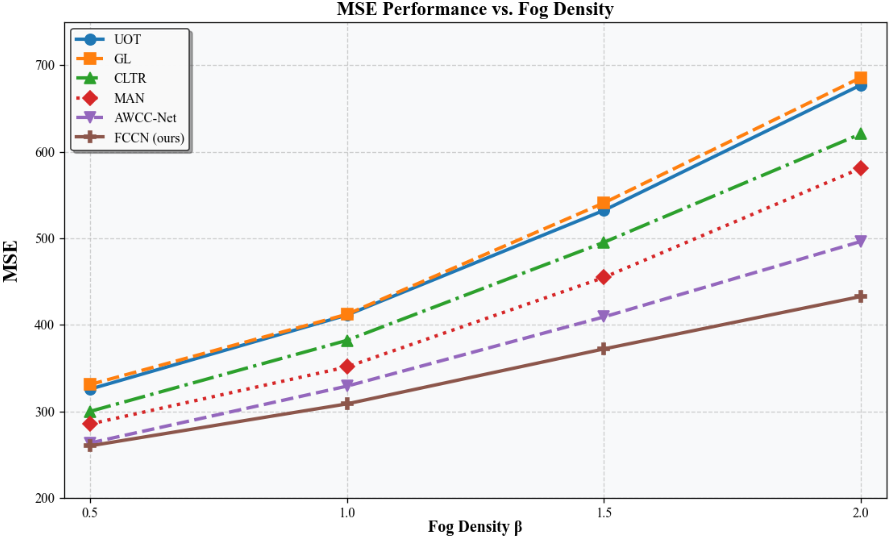}%
\label{fig_second_case}}
\caption{Trends in MAE/MSE for selected advanced methods under different haze concentration conditions. (a) MAE, (b) MSE.}
\label{fig_3}
\vspace{-4mm}
\end{figure}

Figure~\ref{fig_3} clearly demonstrates that fog density significantly impacts the performance of various methods. As haze concentration increases, both MAE and MSE values show an upward trend across all methods. Although performance declines across all approaches, distinct differences emerge in their interference resistance under foggy conditions. At low fog densities ($\beta$ \textless 1.0), the performance gaps between methods are small, with similar MAE and MSE values. However, when $\beta$ exceeds 1.0, the error curves of some traditional methods steeply rise, while the FCCN curve remains relatively flat. For instance, when $\beta$ increased from 1.0 to 2.0, the MAE of traditional methods generally surged by over 35\%, whereas the optimized method's MAE increase was only 22\%, with its advantage becoming more pronounced as fog density rose.
Crucially, the optimized method exhibits minimal error fluctuation across the entire fog density range, with significantly lower standard deviations for both MAE and MSE compared to other comparison methods. This indicates that the method delivers more stable and reliable counting results in complex foggy conditions. Particularly under extreme high fog density ($\beta$ \textgreater 1.5), its error values are approximately 28\% lower than the closest comparison method, fully demonstrating the effectiveness of its fusion-based physical prior approach in extracting robust features.

\subsection{Ablation study}

To gain a deeper understanding of the contribution of each component within the FCCN framework, we designed comprehensive ablation experiments, as shown in Table~\ref{tbl:table3}.

\noindent \textbf{Core value of the Physics-guided module.} After removing the Physics-guided module, MAE and MSE increased significantly by 34.6\% and 30.1\%, respectively. This demonstrates that the Physics-guided module effectively mitigates feature degradation in foggy conditions by modeling fog attenuation effects at the physical mechanism level.

\noindent \textbf{The critical role of MSA-KAN in modeling nonlinear features.} Replacing MSA-KAN with traditional convolutional layers resulted in a 15.5\%/18.7\% decline in model performance and a 21.1\% increase in nonlinear fitting error (measuring deviation between features and true population distribution). This result validates MSA-KAN's advantage in flexibly adapting to the complex nonlinear attenuation patterns of fog-day features.

\noindent \textbf{The necessity of weather-dynamics-aware GCN modules for modeling spatial correlations.} Replacing traditional GCN with weather-dynamics-aware GCN resulted in a 21.3\%/25.5\% increase in MAE/MSE for foggy scenarios. This demonstrates that the dynamic adjacency matrix of weather-dynamics-aware GCN accurately captures the dynamic spatial correlations of crowds in foggy conditions: focusing on local connections during light fog and enabling long-range information complementarity during heavy fog.

\begin{table}[h]
\centering
\vspace{0.3cm}
\begin{tabular}{lcccc}
\hline
\toprule \textbf{Methods} & \textbf{MAE}    & \textbf{MSE}    \\
\midrule
    FCCN (Full)    & \textbf{85.8} &\textbf{421.7} \\
    w/o Phyguided  & 115.5 (+34.6\%) & 548.6 (+30.1\%)  \\
    w/o KAN        & 99.1 (+15.5\%) & 500.1 (+18.7\%)  \\
    w/o WGCN       & 104.1 (+21.3\%) & 529.2 (25.5\%) \\
 
\hline
\end{tabular}
\caption{Main module ablation experiment under foggy conditions. Specifically, variants after removing specific modules.}
\label{tbl:table3}
\end{table}

\begin{figure}[t]
\centering
\subfloat[]{\includegraphics[width=0.5\columnwidth]{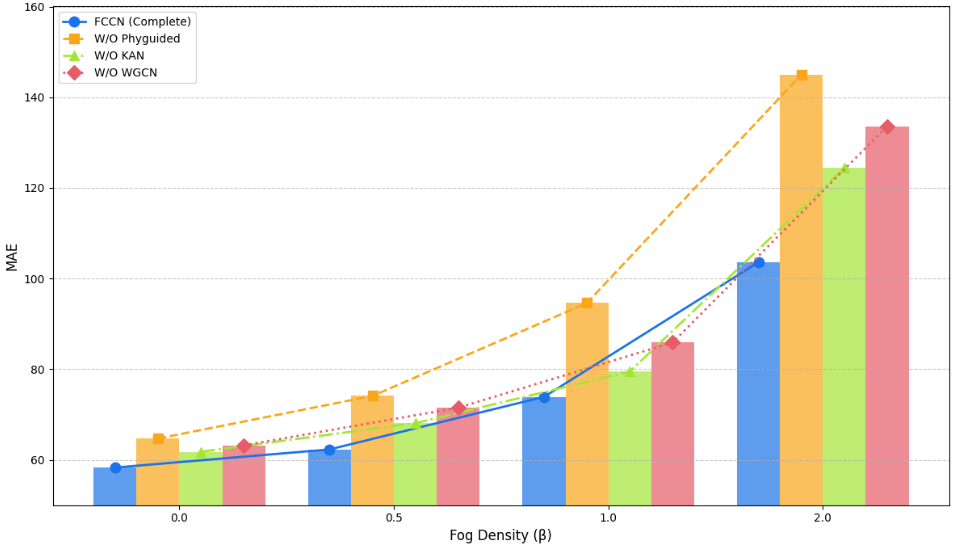}%
\label{fig_first_case}}
\hfil
\subfloat[]{\includegraphics[width=0.5\columnwidth]{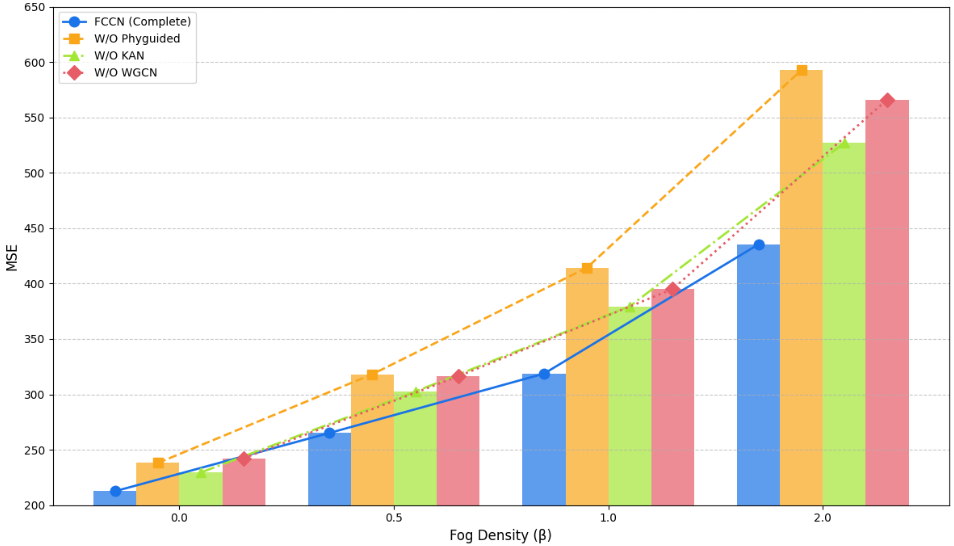}%
\label{fig_second_case}}
\caption{Performance variation curves of FCCN variants under different haze concentrations.
(a) MAE versus haze concentration;
(b) MSE versus haze concentration.}
\label{fig_4}
\vspace{-4mm}
\end{figure}

Figure~\ref{fig_4} provides a comprehensive visualization illustrating how the performance of FCCN and its variants varies at different haze concentration levels ($\beta = 0, 0.5, 1.0, \text{and } 2.0$).

Regarding MAE performance across varying haze concentrations, the curve clearly demonstrates that MAE exhibits an overall upward trend as haze concentration $\beta$ increases. However, the performance gap between the complete FWCC model and its ablation variants continues to widen with rising $\beta$, highlighting its core advantage. The complete model maintains the lowest MAE throughout the haze concentration range with gradual degradation (58.3 at $\beta = 0$, 103.6 at $\beta = 2.0$, representing an increase of 77. 8\%). Conversely, ablation variants degraded more severely: the "w/o Phyguided" variant experienced a steep performance drop of approximately 125.6\% (from 64.3 to 145.0), widening its gap with the full model from 10.2\% at $\beta=0$ to 40\% at $\beta=2.0$. The "W/O KAN" variant degraded by 101.2\%; the "W/O WGCN" variant, constrained by its fixed graph structure, achieved an MAE of 133.6 at $\beta=2.0$, both significantly weaker than the full model.

Analysis of MSE performance across varying fog concentrations demonstrates the robustness of each variant in maintaining counting accuracy. The complete FCCN model exhibits a relatively stable MSE growth trend across all fog density levels, demonstrating outstanding robustness. However, the MSE of the "w/o Phyguided" variant surged sharply from approximately 238.1 at $\beta = 0$ to 592.6 at $\beta = 2.0$ (a 148.9\% increase), widening the gap with the complete model from 12\% to approximately 37.5\%. This highlights the critical role of physical priors in maintaining count stability under high haze conditions. The "w/o KAN" variant exhibits a moderate MSE increase, rising 22.9\% above the full model; the "w/o WGCN" variant shows moderate degradation (133.5\% increase), demonstrating the value of dynamic graph structures in compensating for blurred region errors and maintaining robustness.

In terms of the overall performance of the model, the results indicate that under different haze conditions, the components of FCCN work together to ensure its excellent crowd counting performance. The complete FCCN model exhibits remarkable resilience, with both MAE and MSE degradation rates significantly lower than those of ablation variants as haze concentration increases. The "w/o Phyguided" variant exhibits the most severe performance degradation, confirming the Physics-guided module's pivotal role in quantifying haze attenuation and stabilizing feature inputs. The MSA-KAN layer's parametric nonlinear modeling demonstrates critical value in adapting to the dynamic degradation of hazy-day features. Meanwhile, the Weather-dynamics-aware GCN highlights the complementary role of dynamic graph structures in compensating for errors in ambiguous regions and maintaining counting robustness. As fog density increases, the widening performance gap between the full model and its ablation variants underscores the necessity of an integrated approach. The synergistic effects of these three core components are pivotal for achieving reliable crowd counting under severe weather conditions.

\section{Conclusion}
This paper proposes a fog-penetrating crowd counting framework named FCCN, which deeply integrates atmospheric scattering physics prior with KAN-GCN. We construct a Physics-guided module that achieves dynamic transmission estimation and adaptive calibration of scattering parameters through a physical parameter estimation network. This enables pixel-level modeling of fog attenuation effects and accurately quantifies the nonlinear attenuation patterns of targets at different depths. We introduce MSA-KAN as the feature extraction backbone, embedding physical prior knowledge by replacing fixed activation functions with a combination of parametric basis functions. This significantly enhances the model's ability to discern degraded features and reduces feature confusion error. We designed a Weather-dynamics-aware GCN to construct a dynamic adjacency matrix. Through dual-constraint weighting based on fog density and spatial distance, coupled with weather-aware attention mechanisms, we modeled multi-granularity correlations among crowd targets, enabling collaborative inference in visibility-restricted areas. Additionally, we established a hierarchical haze weather test benchmark to provide evaluation standards for related research. This study offers valuable insights into integrating physical mechanisms with deep learning for visual tasks, while providing reliable support for precise decision-making under complex meteorological conditions in fields such as intelligent monitoring and urban management.

\bibliographystyle{ieee_fullname}
\bibliography{main}

\begin{thebibliography}{10}\itemsep=-1pt

\bibitem{gao2019pccnetperspectivecrowd}
Junyu Gao, Qi Wang, and Xuelong Li.
\newblock Pcc net: Perspective crowd counting via spatial convolutional network, 2019.

\bibitem{7780459}
Kaiming He, Xiangyu Zhang, Shaoqing Ren, and Jian Sun.
\newblock Deep residual learning for image recognition.
\newblock In {\em 2016 IEEE Conference on Computer Vision and Pattern Recognition (CVPR)}, pages 770--778, 2016.

\bibitem{HORNIK1989359}
Kurt Hornik, Maxwell Stinchcombe, and Halbert White.
\newblock Multilayer feedforward networks are universal approximators.
\newblock {\em Neural Networks}, 2(5):359--366, 1989.

\bibitem{huang2023countingcrowdsbadweather}
Zhi-Kai Huang, Wei-Ting Chen, Yuan-Chun Chiang, Sy-Yen Kuo, and Ming-Hsuan Yang.
\newblock Counting crowds in bad weather, 2023.

\bibitem{Idrees2018Composition}
Haroon Idrees, Muhammad Tayyab, Kishan Athrey, Dong Zhang, Somaya Al-Maadeed, Nasir Rajpoot, and Mubarak Shah.
\newblock Composition loss for counting, density map estimation and localization in dense crowds.
\newblock In {\em 2018 European Conference on Computer Vision (ECCV)}, pages 1--16, 2018.

\bibitem{9156690}
Xiaoheng Jiang, Li Zhang, Mingliang Xu, Tianzhu Zhang, Pei Lv, Bing Zhou, Xin Yang, and Yanwei Pang.
\newblock Attention scaling for crowd counting.
\newblock In {\em 2020 IEEE/CVF Conference on Computer Vision and Pattern Recognition (CVPR)}, pages 4705--4714, 2020.

\bibitem{9879174}
Yi Li, Yi Chang, Yan Gao, Changfeng Yu, and Luxin Yan.
\newblock Physically disentangled intra- and inter-domain adaptation for varicolored haze removal.
\newblock In {\em 2022 IEEE/CVF Conference on Computer Vision and Pattern Recognition (CVPR)}, pages 5831--5840, 2022.

\bibitem{8578218}
Yuhong Li, Xiaofan Zhang, and Deming Chen.
\newblock Csrnet: Dilated convolutional neural networks for understanding the highly congested scenes.
\newblock In {\em 2018 IEEE/CVF Conference on Computer Vision and Pattern Recognition}, pages 1091--1100, 2018.

\bibitem{Liang_2022}
Dingkang Liang, Xiwu Chen, Wei Xu, Yu Zhou, and Xiang Bai.
\newblock Transcrowd: weakly-supervised crowd counting with transformers.
\newblock {\em Science China Information Sciences}, 65(6), Apr. 2022.

\bibitem{liang2022endtoendtransformermodelcrowd}
Dingkang Liang, Wei Xu, and Xiang Bai.
\newblock An end-to-end transformer model for crowd localization, 2022.

\bibitem{9880163}
Hui Lin, Zhiheng Ma, Rongrong Ji, Yaowei Wang, and Xiaopeng Hong.
\newblock Boosting crowd counting via multifaceted attention.
\newblock In {\em 2022 IEEE/CVF Conference on Computer Vision and Pattern Recognition (CVPR)}, pages 19596--19605, 2022.

\bibitem{8953396}
Chenchen Liu, Xinyu Weng, and Yadong Mu.
\newblock Recurrent attentive zooming for joint crowd counting and precise localization.
\newblock In {\em 2019 IEEE/CVF Conference on Computer Vision and Pattern Recognition (CVPR)}, pages 1217--1226, 2019.

\bibitem{9010659}
Xiaohong Liu, Yongrui Ma, Zhihao Shi, and Jun Chen.
\newblock Griddehazenet: Attention-based multi-scale network for image dehazing.
\newblock In {\em 2019 IEEE/CVF International Conference on Computer Vision (ICCV)}, pages 7313--7322, 2019.

\bibitem{liu2025kankolmogorovarnoldnetworks}
Ziming Liu, Yixuan Wang, Sachin Vaidya, Fabian Ruehle, James Halverson, Marin Soljačić, Thomas~Y. Hou, and Max Tegmark.
\newblock Kan: Kolmogorov-arnold networks, 2025.

\bibitem{ma2019bayesianlosscrowdcount}
Zhiheng Ma, Xing Wei, Xiaopeng Hong, and Yihong Gong.
\newblock Bayesian loss for crowd count estimation with point supervision, 2019.

\bibitem{Ma2021Learning}
Zhiheng Ma, Xing Wei, Xiaopeng Hong, Hui Lin, Yunfeng Qiu, and Yihong Gong.
\newblock Learning to count via unbalanced optimal transport.
\newblock In {\em 2021 AAAI Conference on Artificial Intelligence (AAAI)}, pages 11224--11232, 2021.

\bibitem{10311083}
Zhuangzhuang Miao, Yong Zhang, Hao Ren, Yongli Hu, and Baocai Yin.
\newblock Multi-level dynamic graph convolutional networks for weakly supervised crowd counting.
\newblock {\em IEEE Transactions on Intelligent Transportation Systems}, 25(5):3483--3495, 2024.

\bibitem{9711067}
Christos Sakaridis, Dengxin Dai, and Luc Van~Gool.
\newblock Acdc: The adverse conditions dataset with correspondences for semantic driving scene understanding.
\newblock In {\em 2021 IEEE/CVF International Conference on Computer Vision (ICCV)}, pages 10745--10755, 2021.

\bibitem{sam2020locatesizecountaccurately}
Deepak~Babu Sam, Skand~Vishwanath Peri, Mukuntha~Narayanan Sundararaman, Amogh Kamath, and R.~Venkatesh Babu.
\newblock Locate, size and count: Accurately resolving people in dense crowds via detection, 2020.

\bibitem{10011779}
Qingzhen Shang, Yidi Zhuo, Jiahui Zhang, and Jinfu Yang.
\newblock Rdnet: A reinforced-deformable convolutional network for crowd counting.
\newblock In {\em 2022 IEEE International Conference on Robotics and Biomimetics (ROBIO)}, pages 1787--1792, 2022.

\bibitem{9248596}
Vishwanath~A. Sindagi, Rajeev Yasarla, and Vishal~M. Patel.
\newblock Jhu-crowd++: Large-scale crowd counting dataset and a benchmark method.
\newblock {\em IEEE Transactions on Pattern Analysis and Machine Intelligence}, 44(5):2594--2609, 2022.

\bibitem{2021Rethinking}
Qingyu Song, Changan Wang, Zhengkai Jiang, Yabiao Wang, and Yang Wu.
\newblock Rethinking counting and localization in crowds:a purely point-based framework.
\newblock 2021.

\bibitem{10076399}
Yuda Song, Zhuqing He, Hui Qian, and Xin Du.
\newblock Vision transformers for single image dehazing.
\newblock {\em IEEE Transactions on Image Processing}, 32:1927--1941, 2023.

\bibitem{Wakaura2025Adaptive}
Hikaru Wakaura, Rahmat Mulyawan, and Andriyan~B. Suksmono.
\newblock Adaptive variational quantum kolmogorov-arnold network.
\newblock {\em arXiv preprint arXiv:2503.21336}, 2025.

\bibitem{9578673}
Jia Wan, Ziquan Liu, and Antoni~B. Chan.
\newblock A generalized loss function for crowd counting and localization.
\newblock In {\em 2021 IEEE/CVF Conference on Computer Vision and Pattern Recognition (CVPR)}, pages 1974--1983, 2021.

\bibitem{wang2020distributionmatchingcrowdcounting}
Boyu Wang, Huidong Liu, Dimitris Samaras, and Minh Hoai.
\newblock Distribution matching for crowd counting, 2020.

\bibitem{9153156}
Qi Wang, Junyu Gao, Wei Lin, and Xuelong Li.
\newblock Nwpu-crowd: A large-scale benchmark for crowd counting and localization.
\newblock {\em IEEE Transactions on Pattern Analysis and Machine Intelligence}, 43(6):2141--2149, 2021.

\bibitem{wang2019learningsyntheticdatacrowd}
Qi Wang, Junyu Gao, Wei Lin, and Yuan Yuan.
\newblock Learning from synthetic data for crowd counting in the wild, 2019.

\bibitem{9578448}
Haiyan Wu, Yanyun Qu, Shaohui Lin, Jian Zhou, Ruizhi Qiao, Zhizhong Zhang, Yuan Xie, and Lizhuang Ma.
\newblock Contrastive learning for compact single image dehazing.
\newblock In {\em 2021 IEEE/CVF Conference on Computer Vision and Pattern Recognition (CVPR)}, pages 10546--10555, 2021.

\bibitem{9859777}
Zhengtao Wu, Lingbo Liu, Yang Zhang, Mingzhi Mao, Liang Lin, and Guanbin Li.
\newblock Multimodal crowd counting with mutual attention transformers.
\newblock In {\em 2022 IEEE International Conference on Multimedia and Expo (ICME)}, pages 1--6, 2022.

\bibitem{7780439}
Yingying Zhang, Desen Zhou, Siqin Chen, Shenghua Gao, and Yi Ma.
\newblock Single-image crowd counting via multi-column convolutional neural network.
\newblock In {\em 2016 IEEE Conference on Computer Vision and Pattern Recognition (CVPR)}, pages 589--597, 2016.

\end{thebibliography}
\vfill

\end{document}